\documentclass[10pt, conference, compsocconf]{IEEEtran}
\usepackage{ifpdf}
\usepackage{cite}
\ifCLASSINFOpdf
\usepackage[pdftex]{graphicx}
\else
\usepackage[dvips]{graphicx}
\fi
\usepackage[cmex10]{amsmath}
\usepackage{caption}
\usepackage{amssymb}
\usepackage{algorithmic}
\usepackage{array}
\usepackage{mdwmath}
\usepackage{mdwtab}
\usepackage{eqparbox}
\usepackage{epsfig}
\usepackage[most]{tcolorbox}
\usepackage[tight,footnotesize]{subfigure}
\usepackage{caption}
\usepackage[font=footnotesize]{subfig}
\usepackage{dblfloatfix}
\usepackage{url}
\usepackage{soul}

\begin{document}

\title{Textual Description for Mathematical Equations}
\author{\IEEEauthorblockN{Ajoy Mondal and C V Jawahar}
\IEEEauthorblockA{Centre for Visual Information Technology,\\
International Institute of Information Technology, Hyderabad, India, \\
Email: ajoy.mondal@iiit.ac.in and jawahar@iiit.ac.in}}
\maketitle

\begin{abstract}
Reading of mathematical expression or equation in the document images is very challenging due to the large variability of mathematical symbols and expressions. In this paper, we pose reading of mathematical equation as a task of generation of the textual description which interprets the internal meaning of this equation. Inspired by the natural image captioning problem in computer vision, we present a mathematical equation description ({\sc \textbf{med}}) model, a novel end-to-end trainable deep neural network based approach that learns to generate a textual description for reading mathematical equation images. Our {\sc \textbf{med}} model consists of a convolution neural network as an encoder that extracts features of input mathematical equation images and a recurrent neural network with attention mechanism which generates description related to the input mathematical equation images. Due to the unavailability of mathematical equation image data sets with their textual descriptions, we generate two data sets for experimental purpose. To validate the effectiveness of our {\sc \textbf{med}} model, we conduct a real-world experiment to see whether the students are able to write equations by only reading or listening their textual descriptions or not. Experiments conclude that the students are able to write most of the equations correctly by reading their textual descriptions only.    
\end{abstract}


\maketitle

\section{Introduction} \label{introduction}

Learning of mathematics is necessary for students at every stage of student life. Solving mathematical problems is an alternative way to develop and improve their mathematical skills. Unfortunately, blind and visually impaired ({\sc vi}) students face the difficulties to especially learn mathematics due to their limitations in reading and writing mathematical formulas. In generally, human readers help those categories of students to access and interpret materials or documents of mathematics courses. However, at every time, it is impossible and impractical for those categories of students having human reader; because of the cost and the limited availability of the trained personnel. Braille is the popular and more convenient way to access the document for blind and {\sc vi} students. Unfortunately, many documents are not available in Braille, since the conversion of mathematical documents in Braille is more difficult and complicated~\cite{math_blind_survey}. Moreover, it is also difficult for students who are comfortable for reading literary Braille transcriptions~\cite{imath_2012}.

Other than Braille, sound based representation of documents is also an important and popular way to access information 
for those kinds of students. In this direction, {\sc daisy} books and talking books are commonly used audio materials 
for understanding documents. However, these books are less prepared for mathematical expressions or equations ({\sc me}s)~\cite{imath_2012}. 
Recently, text-to-speech ({\sc tts}) systems have been widely used by blind and {\sc vi} students to read electronic text through computers. {\sc tts} systems convert digital text into synthetic speech. Unfortunately, most available {\sc tts} systems can read only plain text. They fail to generate appropriate speech when it comes across mathematical equations.

Many researchers realized that it is very important to enhance the accessibility of mathematical materials to the blind and {\sc vi} students and developed {\sc tts} based mathematical expressions reading systems~\cite{soiffer2005mathplayer,math_audio,medjkoune2017combining}. Some of these developed systems need extra words to read an {\sc me}. Due to the extra words, the rendered audio is very long. Hence, the students may not always be able to interpret the main point of the expressions due to long audio duration. Moreover, most of these existing automatic math reading systems take {\sc me}s as input in the form of LaTeX or other similar markup languages. Unfortunately, all available mathematical documents are not in the form of LaTeX or any other markup language. Therefore, generation of LaTeX or other markup languages corresponds to mathematical documents is also challenging~\cite{imath_2012}.   

\begin{figure}[t]
\centerline{
\psfig{figure=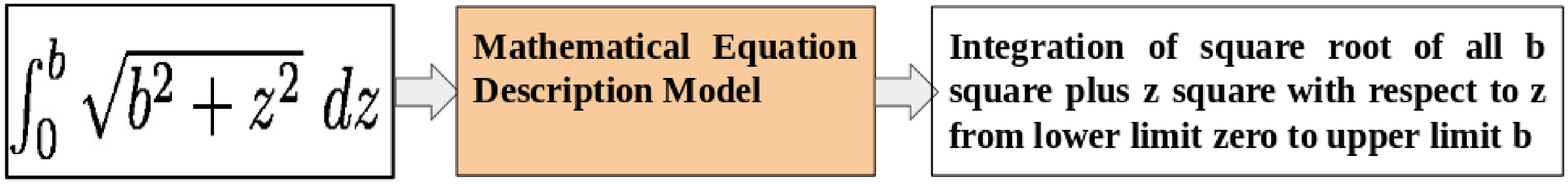,height = 0.05\textwidth, width=0.5\textwidth}
\hspace{0.0001\textwidth}}
\caption{Our model treats reading of mathematical equation in a document image as generation of textual description which interprets the internal meaning of this equation.}\label{fig_intro}
\end{figure}

\begin{figure*}[ht!]
\centerline{
\psfig{figure=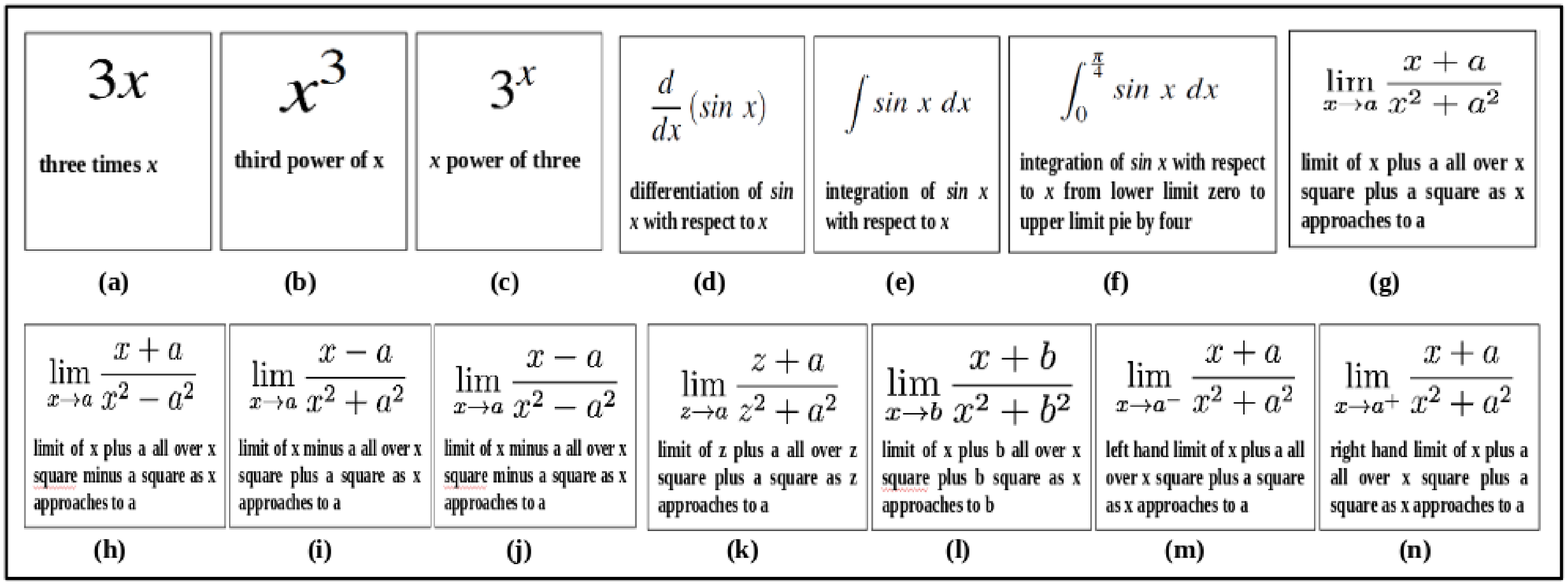,height = 0.21\textwidth, width=1.0\textwidth}
\hspace{0.0001\textwidth}}
\caption{Example of sensitivity of variables, operators and their positions while reading the equations. Only `$3$' in (a) is changing position in (b), `$x$' in (a) is changing position in (c), `differentiation operator' in (d) is changed by `integration' in (e), and `differentiation operator' in (d) is changed by `finite integration' in (f), `$+$' operator in denominator of (g) is changed by `$-$' in (h), `$+$' operator in nominator of (g) is changed by `$-$' in (i), `$+$' operators in both nominator and denominator of (g) are changed by `$-$' in (j), variable `$x$' in (g) is changed by `$z$' in (k), constant `$a$' in (g) is changed by `$b$' in (l), limit value `$a$' in (g) is replaced by limit value `$a^{-}$' in (m), and limit value `$a$' in (g) is changed to `$a^{+}$'in (n). \label{fig_sensitivity}}
\end{figure*}

In this paper, our goal is to develop a framework, called as mathematical equation description (\textsc{med}) model which can help the blind and {\sc vi} students for reading/interpreting internal meaning of {\sc me}s present in the document images. We pose reading of {\sc me} as a problem of generation of natural language description. Basically, our proposed \textsc{med} model automatically generates textual (natural language) description which can able to interpret the internal meaning of the {\sc me}. For examples, $\int sin\ x\ dx$ is a {\sc me} and its textual description is like ``integration of $sin$ $x$ with respect to $x$". With the textual description, the blind and {\sc vi} students can easily read/interpret the {\sc me}s. Figure~\ref{fig_intro} shows the generated textual description of an \textsc{mei} using our \textsc{med} model. This task is closely related to image description/captioning task~\cite{karpathy2015deep}. However, description of equation is very sensitive with respect to variables, operators and their positions. Figure~\ref{fig_sensitivity} illustrates the sensitivity of variables, operators and their positions during generation of textual description for \textsc{me}s. For example, $3x$ in Figure~\ref{fig_sensitivity}(a), it can be read as ``three times $x$". While $3$ is changing its position e.g. $3^{x}$ in Figure~\ref{fig_sensitivity}(c), textual sentence ``$x$ power of three" for reading, it is totally different from previous equation. As per our understanding goes, this is the first work where reading/interpreting of {\sc me}s is posed as a generation of textual description task.  

The main inspiration of our work comes from image captioning, a recent advancement in computer vision. In this paper, we propose an end-to-end trainable deep network to 
generate natural language descriptions for the {\sc me}s which can read/interpret the internal meaning of these expressions. The network consists of two modules: encoder and decoder. The encoder encodes the {\sc me} images using Convolution
Neural Networks ({\sc cnn}s). Long Short Term Memory ({\sc lstm}) network as decoder that takes the intermediate representation to generate textual descriptions of corresponding {\sc me} images. The attention mechanism impels the decoder to focus on specific parts of the input image. The encoder, decoder and attention
mechanism are trained in a joint manner. We refer this network as Mathematical Equation Description ({\sc med}) network. 

In particular, our contributions are as follows.

\begin{itemize}
\item We present an end-to-end trainable deep network with the combination of both vision and language models to generate description of {\sc me}s for reading/interpreting {\sc me}s .

\item We generate two data sets with {\sc me} images and their corresponding natural language descriptions for our experimental purpose.

\item We conduct a real world experiment to establish effectiveness of the {\sc med} model for reading/interpreting mathematical equation.
\end{itemize}

\begin{figure*}[ht!]
\centerline{
\includegraphics[width=18cm,height=3.5cm]{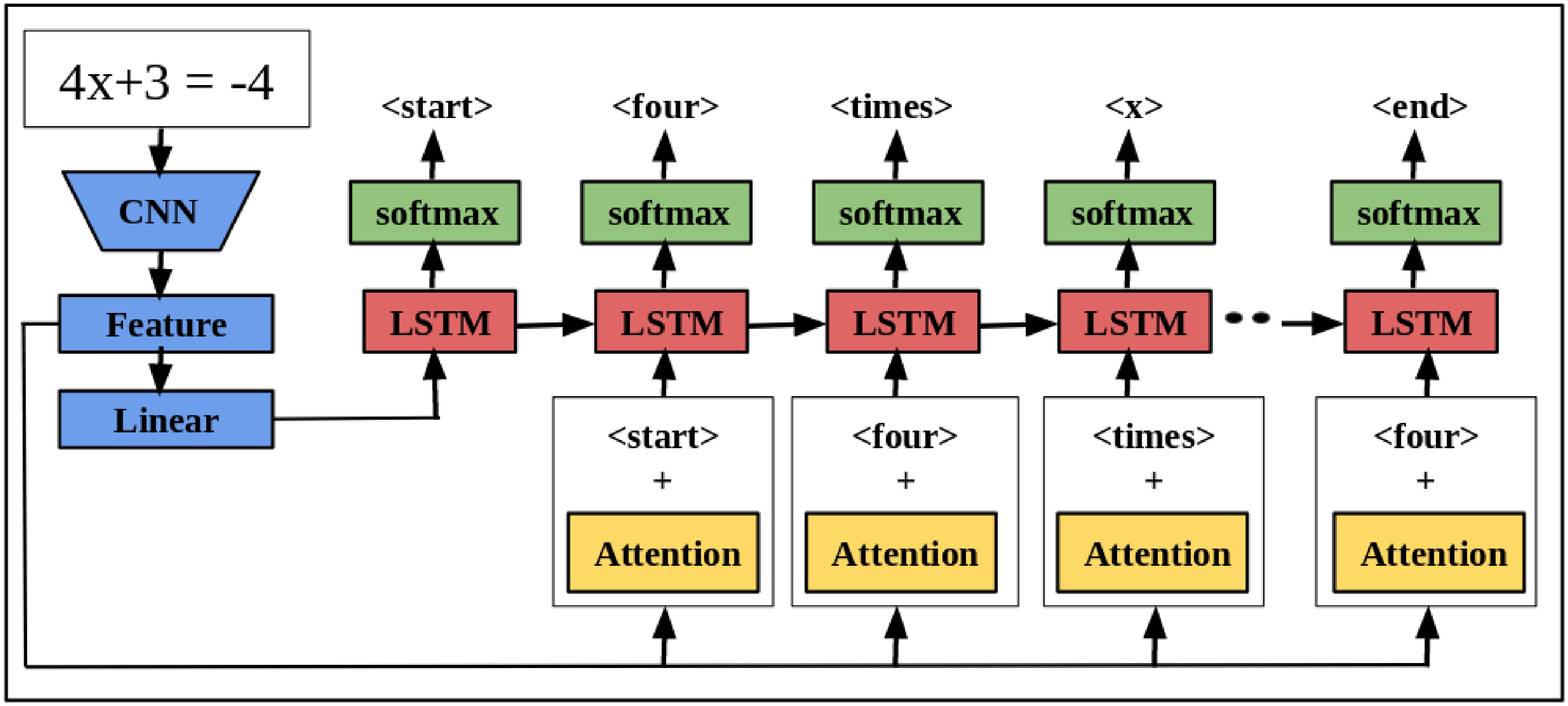}}
\caption{Overview of mathematical expression description network. Our model uses a end-to-end trainable network consisting of {\sc cnn} followed by a language generating {\sc lstm}. It generates textual description of an input mathematical expression image in natural language which interprets its internal meaning.\label{figure_model}}
\end{figure*} 

\section{Related Work} \label{related_work}

\subsection{Mathematical Expression Recognition}\label{mathematical_expression_recognition}

Automatic recognition of {\sc me}s is one of the major tasks towards transcribing documents into digital form in the scientific and engineering fields. This task mainly consists of two major steps: symbol recognition and structural analysis~\cite{chan2000mathematical}. In case of symbols recognition, the initial task is to segment the symbols and then to recognize the segmented symbols. Finally, structural analysis of the recognized symbols have been done to recognize the mathematical expressions. These two problems can be solved either sequentially~\cite{zanibbi2002recognizing} or a single framework (global)~\cite{alvaro2016integrated}. However, both of these sequential and global approaches have several limitations including (i) segmentation of mathematical symbols is challenging for both printed and handwritten documents as it contains a mix of text, expressions and figures; (ii)
symbols recognition is difficult because a
large number of symbols, fonts, typefaces and font sizes~\cite{chan2000mathematical}; (iii) for structural analysis, commonly two dimensional context free grammar is used which requires prior knowledge to define math grammar~\cite{alvaro2011recognition} and (iv)
the complexity of parsing algorithm increases with the size of math grammar~\cite{julca2015top}.

Due to the success of deep neural network in computer vision task, the researchers adopted deep neural 
models to recognize mathematical symbols~\cite{dai2015deep,dai2016recognition} and expressions~\cite{DengKR16,deng2017image,zhang2018multi,zhang2017gru,ZHANG2017196}.   
In~\cite{dai2015deep,dai2016recognition}, the authors considered {\sc cnn} along with bidirectional {\sc lstm} to recognize mathematical symbols.
Whereas,~\cite{DengKR16,deng2017image} explored the use of attention based image-to-text models for generating structured markup for {\sc me}s. These models consist of a multi-layer convolution network to extract image features with an attention based recurrent neural network as a decoder for generating structured markup text. In the same direction, Zhang {\em et al.}~\cite{ZHANG2017196} proposed a novel end-to-end approach based on neural network that learns to recognize handwritten mathematical expressions ({\sc hme}s) in a two-dimensional layout and produces output as one-dimensional character sequences in the LaTeX format. Here, the {\sc cnn}, as encoder is considered to extract feature from {\sc hme}s images and a recurrent neural network is employed as decoder to generate LaTeX sequences. 

\subsection{Image Captioning} \label{image_captioning}

Image captioning is a task which automatically describes the content of an image using properly formed English sentences. Although, it is a very challenging task, it helps the visually impaired people to better understand the content of images on the web. Recently, a large variety of deep models~\cite{chen2015mind,donahue2015long,karpathy2015deep,vinyals2015show,you2016image} have been proposed to generate textual description of natural images. All these models considered recurrent neural network ({\sc rnn}) as language models conditioned on the image features extracted by convolution neural networks and sample from them to generate text. Instead of generating caption for whole image, a handful of approaches to generate captions for image regions~\cite{xu2015show,karpathy2015deep,johnson2016densecap}. In contrast of generating a sentence, various models have also been been introduced to generate paragraph for describing content of the images in literature~\cite{yu2016video,krause2016hierarchical} by considering a hierarchy of language models.

\section{Mathematical Equation Description}

\subsection{Overview}

Our {\sc med} model takes a mathematical expression image ({\sc mei}) as an input and generates a natural language sentence to describe the internal meaning of this expression. Figure~\ref{figure_model} provides an overview of our model. It consists of encoder and decoder networks. The encoder extracts deep features to richly represent the equation images. The decoder uses the intermediate representation to generate a sentence to describe the meaning of the {\sc me}. The attention mechanism impels the decoder to focus on specific parts of the input image. Each of these networks are discussed in details in the following subsections. 

\subsection{Feature Extraction using Encoder}

The {\sc med} model takes a {\sc mei} and generates its textual description $\mathbf{Y}$ encoded as a sequence of $1$-of-$K$ encoded words.

\begin{equation}
\mathbf{Y} = \{\mathbf{y}_1, \mathbf{y}_2, ..., \mathbf{y}_T   \},\ \mathbf{y}_i \in \mathbb{R}^K
\end{equation}
where $K$ is the size of the vocabulary and $T$ is the length of the description. We consider a Convolution Neural Network ({\sc cnn}) as an encoder in order to extract a set of feature vectors. We assume that the output of {\sc cnn} encoder is a three-dimensional array of size $H \times W \times D$, and consider the output as a variable length grid of $L$ vectors, $L = H \times W$ as referred to annotation vectors. Each of these vector is $D$-dimensional representation that corresponds to a local region of the input image.
\begin{equation}
\mathbf{A} = \{\mathbf{a}_1, \mathbf{a}_2,...,\mathbf{a}_L\},\ \mathbf{a}_i \in \mathbb{R}^{D}    
\end{equation}

We extract features from a lower convolution layer in order to obtain a correspondence between the feature vectors and regions of the image. This allows the decoder to selectively focus on certain regions of the input image by selecting a subset of all these feature vectors.

\subsection{Sentence Generation using Decoder}

We employ {\sc lstm}~\cite{hochreiter1997long} as a decoder that produces a sentence by generating one word at every time step conditioned on a context vector $\mathbf{\hat{z}}_{t}$, the hidden state $\mathbf{h}_{t}$ and the previously generated word $\mathbf{y}_{t-1}$. It produces word at time step $t$ using the following equation:

\begin{equation}
p(\mathbf{y}_{t}|\mathbf{y}_{1}, \mathbf{y}_{2},..., \mathbf{y}_{t-1}, \mathbf{x}) = f(\mathbf{y}_{t-1}, \mathbf{h}_{t}, \mathbf{\hat{z}}_{t}), \label{equation_probability}    
\end{equation}
where $\mathbf{x}$ denotes the input {\sc mei} and $f$ denotes a multi-layered perceptron ({\sc mlp}) which is expanded in Eq.~(\ref{equation_final_probability}). The hidden state $\mathbf{h}_t$ of {\sc lstm} is computed using following equation:

\begin{equation}
\begin{array}{l}
\mathbf{i}_{t} = \sigma(\mathbf{W}_{yi} \mathbf{Ey}_{t-1} + \mathbf{U}_{hi} \mathbf{h}_{t-1} + \mathbf{V}_{zi} \mathbf{\hat{z}}_{t}) \\
\mathbf{f}_{t} = \sigma(\mathbf{W}_{yf} \mathbf{Ey}_{t-1} + \mathbf{U}_{hf} \mathbf{h}_{t-1} + \mathbf{V}_{zf} \mathbf{\hat{z}}_{t}) \\
\mathbf{o}_{t} = \sigma(\mathbf{W}_{yo} \mathbf{Ey}_{t-1} + \mathbf{U}_{ho} \mathbf{h}_{t-1} + \mathbf{V}_{zo} \mathbf{\hat{z}}_{t})\\
\mathbf{g}_{t} = \tanh((\mathbf{W}_{yc} \mathbf{Ey}_{t-1} + \mathbf{U}_{hc} \mathbf{h}_{t-1} + \mathbf{V}_{zc} \mathbf{\hat{z}}_{t})\\
\mathbf{c}_{t} = \mathbf{f}_{t} \odot \mathbf{c}_{t-1} + \mathbf{i}_{t} \odot \mathbf{g}_{t} \\
\mathbf{h}_{t} = \mathbf{o}_{t} \odot \tanh(\mathbf{c}_{t}) \label{equn_fun1}.
\end{array}
\end{equation}

Here, $\mathbf{i}_{t}$, $\mathbf{f}_{t}$, $\mathbf{c}_{t}$, $\mathbf{o}_{t}$ and $\mathbf{h}_{t}$
are the input, forget, memory, output and hidden states of the {\sc lstm}, respectively. The vector $\mathbf{\hat{z}}_{t}$ is a context vector which captures visual information of a particular image region. The context vector $\mathbf{\hat{z}}_{t}$ (in Eq.~(\ref{equn_fun1})) is a dynamic representation of the relevant part of the input image at time step $t$. We consider soft attention defined by Bahdannu {\em et al.}~\cite{bahdanau2014neural} which computes weight $\alpha_{ti}$ of each annotation vectors $\mathbf{a}_{i}$ conditioned on the previous {\sc lstm} hidden state $\mathbf{h}_{t-1}$. Here, we parameterize attention as {\sc mlp} which is jointly trained:

\begin{equation}
\begin{array}{l}
e_{ti}=\mathbf{v}_{a}^{T} \tanh(\mathbf{W}_{a} \mathbf{h}_{t} + \mathbf{U}_{a} \mathbf{a}_{i}) \\     
\alpha_{ti} = \frac{exp(e_{ti})}{\sum_{k=1}^{L} exp(e_{tk})}. 
\end{array}
\end{equation}

Let $n^{'}$ be the dimension of the attention, then $\mathbf{v}_{a}$ $\in$ $\mathbb{R}^{n^{'} \times n}$, $\mathbf{U}_{a}$ $\in$ $\mathbb{R}^{n^{'} \times D}$. After computation of the weights $\alpha_{ti}$, the context vector $\mathbf{\hat{z}}_{t}$ is calculated as follows:

\begin{equation}
\mathbf{\hat{z}}_{t} = \sum_{i=1}^{L} \alpha_{ti} \mathbf{a}_{i}.    
\end{equation}

This weight $\alpha_{ti}$ will make decoder to know which part of input image is the suitable place to attend to generate the next predicted word and then assign a higher weight to the corresponding annotation vectors $\mathbf{a}_{i}$. $m$ and $n$ denote the dimensions of embedding and {\sc lstm}, respectively; $E \in \mathbb{R}^{m \times K}$ is the embedding matrix. $\sigma$ is sigmoid activation function and $\odot$ is element wise multiplication.  

Finally, the probability of each predicted word at time $t$ is computed by the context vector $\mathbf{\hat{z}}_{t}$, current {\sc lstm} hidden state $\mathbf{h}_{t}$ and previous predicted word $\mathbf{y}_{t-1}$ using the following equation:

\begin{equation}
\begin{split}
p(\mathbf{y}_{t}|\mathbf{y}_{1}, \mathbf{y}_{2},..., \mathbf{y}_{t-1},\ \mathbf{x}) = \\ g(\mathbf{W}_{o}(\mathbf{E}\mathbf{y}_{t-1} + \mathbf{W}_{h} \mathbf{h}_{t}+ \mathbf{W}_{z} \mathbf{\hat{z}}_{t})), \label{equation_final_probability}
\end{split}
\end{equation}

where $g$ denotes a softmax activation function over all the words in the vocabulary; $\mathbf{E}$, $\mathbf{W}_{o}$ $\in$ $\mathbb{R}^{K \times n}$, $\mathbf{W}_{h}$ $\in$ $\mathbb{R}^{m \times n}$ and $\mathbf{W}_{c}$ $\in$ $\mathbb{R}^{m \times D}$ are learned parameters initialized randomly.

The initial memory state $\mathbf{c}_{0}$ and hidden state $\mathbf{h}_{0}$ of the {\sc lstm} are predicted by an average of the annotation vectors fed through two separate {\sc mlp}s ($f_{init,\ c}$, $f_{init,\ h}$)

\begin{equation}
\begin{array}{l}
\mathbf{c}_{0} = f_{init,\ c} \left(\frac{1}{L} \sum_{i=1}^{L} \mathbf{a}_{i} \right)  \\  
\mathbf{h}_{0} = f_{init,\ h} \left(\frac{1}{L}\sum_{i=1}^{L} \mathbf{a}_{i} \right). 
\end{array}   
\end{equation}

\begin{table}[ht!]
\addtolength{\tabcolsep}{-0.8pt}
\begin{center}
\begin{tabular}{|l|l|} \hline
\textbf{Expression} & \textbf{Description} \\ \hline
$10x$      & ten times $x$ \\ 
$x^{2}$    & $x$ square or second power of $x$ \\ 
$\sqrt{x}$ & second root of $x$    \\ 
$\frac{x}{10}$ & $x$ over ten       \\ 
$(x+y)^{2}$ & second power of all $x$ plus $y$ \\ 
$log_{2}x$  & log $x$ to base two \\ 
$\frac{x}{y}$ & $x$ over $y$      \\ 
$\frac{x}{y+z}$ & $x$ over $y$ plus $z$ \\ 
$\frac{x+y}{z}$ & $x$ plus $y$ all over $z$ \\ 
$\frac{x}{y^{2}}$ & $x$ over $y$ square \\ 
$x+\frac{y}{z}$   & $x$ all plus $y$ over $z$ \\ 
$\frac{x-y}{y+z}$ & $x$ minus $y$ all over $y$ plus $z$ \\ 
$\frac{x^{2}}{y}$ & $x$ square over $y$ \\ 
$\frac{x-y}{z}+t$ & $x$ minus $y$ all over $z$ all plus $t$ \\ 
$e^{(1+x)}$ & exponential of all one plus $x$  \\ 
$e^{x}+1$ & exponential of $x$ all plus one \\ 
$e^{(1+x)}-1$ & exponential of all one plus $x$ all minus one \\ 
$\int x\ dx$  & integral of $x$ with respect to $x$ \\
$\displaystyle \int_{0}^{1} x \ dx$ & integral of $x$ with respect to $x$ from lower limit \\ 
 & zero to upper limit one\\ 
$\displaystyle \lim_{x\to 0}\frac{sin\ x}{x}$ & limit of $sin$ $x$ over $x$ as $x$ approaches to zero \\
$\frac{d}{dx}(x^{2})$ & differentiation of $x$ square with respect to $x$ \\
$x-6 = 3$       & $x$ minus six equal to three \\
$x+7>10$        & $x$ plus seven greater than ten \\
$x+2y=7$,  & $x$ plus two times $y$ equal to seven and \\
$x-y=3$    & $x$ minus $y$ equal to three \\ \hline
\end{tabular}
\end{center}
\caption{Natural language phrases to uniquely describe mathematical equations.}\label{table_NLT}
\end{table}

\subsection{Implementation Details}

\paragraph{Pre-processing}

Textual descriptions of the {\sc me}s are pre-processed with basic tokenization algorithm by keeping all words that appeared at least $4$ times in the training set.

\begin{figure*}[ht!]
\centerline{
\psfig{figure=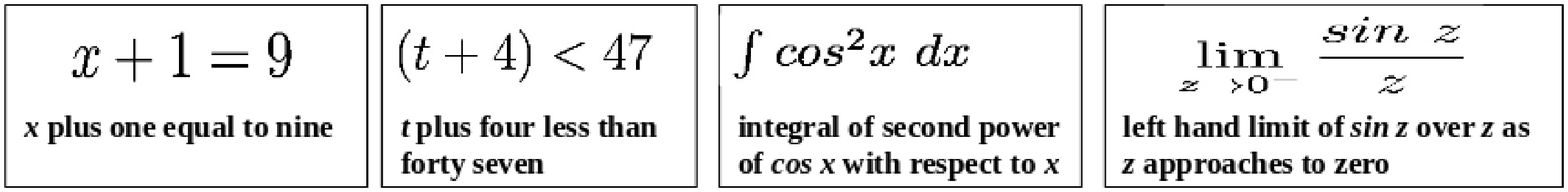,height =0.1\textwidth, width=0.5\textwidth}
\hspace{-0.01\textwidth}
\psfig{figure=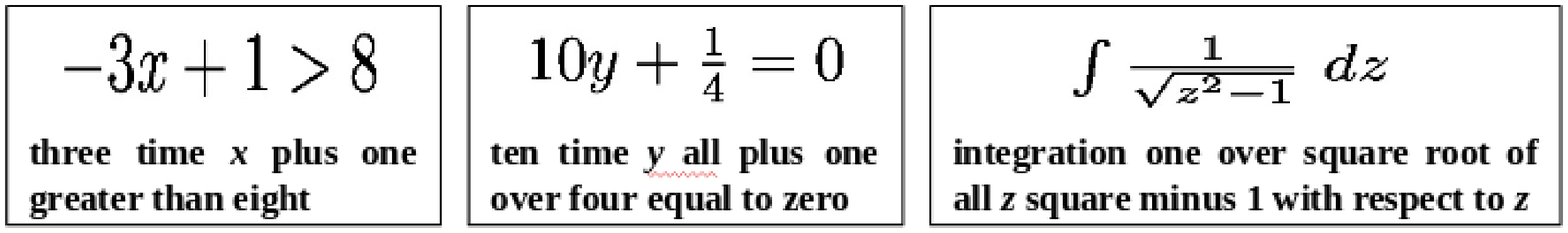,height =0.1\textwidth, width=0.5\textwidth}}
\vspace{0.0001\textwidth}
\centerline{
\psfig{figure=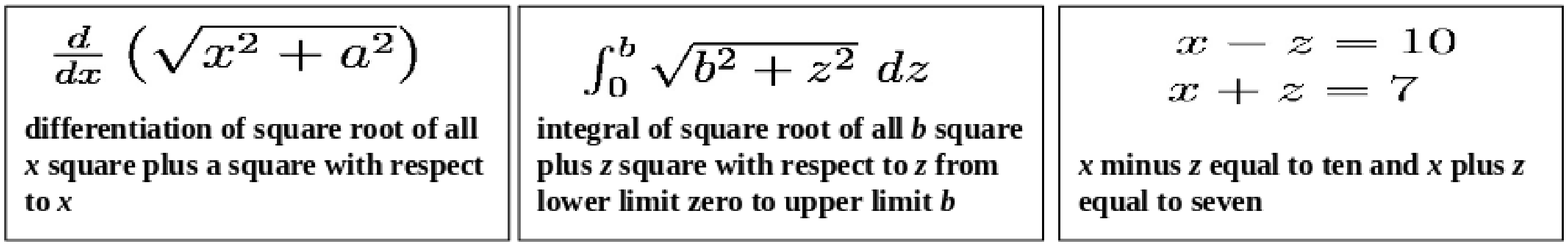,height =0.1\textwidth, width=0.5\textwidth}
\hspace{-0.01\textwidth}
\psfig{figure=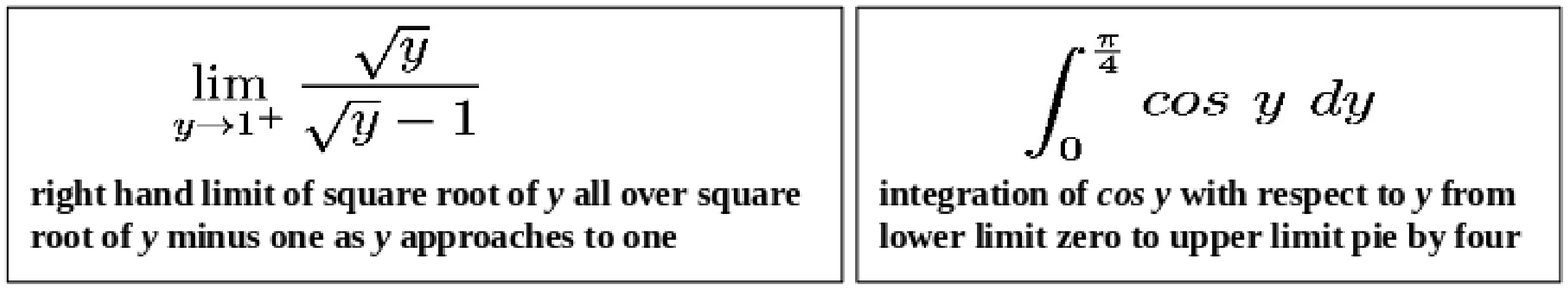,height =0.1\textwidth, width=0.5\textwidth}}
\caption{Few sample {\sc mei} and their corresponding textual description of \texttt{Math-Exp-Syn} data set. \label{fig_sample_img1}}
\end{figure*}

\paragraph{Training Details}

The training objective of our {\sc med} model is to maximize the predicted word probability as given in Eq.~(\ref{equation_final_probability}). We use cross entropy as the objective function:
\begin{equation}
O = -\sum_{t=1}^{T} \log p(y_{t}^{gt}|\mathbf{y}_{t},\ \mathbf{x}),     
\end{equation}
where $y_{t}^{gt}$ represents the ground truth word at time step $t$.

We consider $299$K images and their corresponding textual descriptions to train the model. We consider pre-trained
ResNet-$152$ model~\cite{he2016deep} (on ImageNet~\cite{imagenet_cvpr09}). We do this in all the experiments. We train the network with batch size of $50$ for $60$ epochs. We use stochastic gradient descent ({\sc sgd}) with fixed learning rate $10^{-4}$, momentum $= 0.5$ and weight decay $= 0.0001$. All the weights were randomly initialized except for the {\sc cnn}. We use $512$ dimensions for the embedding and $1024$ for the size of the {\sc lstm} memory. We consider a dropout layer after each convolution layer and set as $0.5$. The best trained model is determined in terms of {\sc bleu}-$4$ score on validation set. For further implementation and architecture details, please refer to the source code at: \textit{\url{https://github.com/ajoymondal/Equation-Description-PyTorch}}.  

\paragraph{Decoding}

In decoding stage, our main aim is to generate a most likely textual description for a given {\sc mei}:
\begin{equation}
\mathbf{\hat{y}} = \operatorname*{arg\,max}_{\mathbf{y}}\ \log p(\mathbf{y}|\mathbf{x}).  
\end{equation}

Different from training procedure, the ground truth of previous predicted word is not available. We employ beam search~\cite{cho2015natural} of size $20$ during decoding procedure. A set of $20$ partial hypothesis beginning with the start-of-sentence $<$start$>$ is maintained. At each time step, each partial hypothesis in the beam is expanded with every possible word. Only the $20$ most likely beams are kept. When the $<$start$>$ token is encounter, it is removed from the beam and added to the set of complete hypothesis. This process is repeated until the output word becomes a symbol corresponding to the end-of-sentence $<$end$>$.

\section{Data Sets and Evaluation Metrics} \label{section_data_set}

\subsection{Data Sets}

Unavailability of mathematical equation image data sets and their textual descriptions inspired us to generate data sets for experimental purpose. Various issues must be concerned during generation of unambiguous textual description of a mathematical equation. One important issue is
that the same textual description can lead to the different expressions. For example, the textual description like
``$x$ plus $y$ over $z$'' could be description of two possible equations: either $\frac{x+y}{z}$ or $x+\frac{y}{z}$.
Thus, an algorithm should be carefully designed to generate an unambiguous textual description corresponds to exactly one expression. As per our knowledge goes, no mathematical expression data sets with their textual descriptions is available for experiment. We create a data set, referred as \texttt{Math-Exp-Syn} with large number of synthetically generated {\sc mei}s and their descriptions. For this purpose, we create sets of predefined functions (e.g. linear equation, limit, etc.), variables (e.g. $x$, $y$, $z$, etc.), operators (e.g. $+$, $-$, etc.) and constants (e.g. $10$, $1$, etc.) and sets of their corresponding textual descriptions. We develop a python code which randomly selects a function, variable, operator and constant from the corresponding predefined sets and automatically generates mathematical equation as an image and corresponding textual description in the text format. We make our \texttt{Math-Exp-Syn} data generation code available at: \textit{\url{https://github.com/ajoymondal/Equation-Description-PyTorch}}. We also create another data set, referred as \texttt{Math-Exp} by manually annotating a limited number of {\sc mei}s. During creation of both these data sets, we take care the uniqueness of the equations and their descriptions. We consider the following natural language sentences listed in table~\ref{table_NLT} to uniquely describe the internal meaning of the equations.

In this work, we limit ourselves to only seven 
categories of {\sc me}s: \textit{linear equation}, \textit{inequality}, \textit{pair of linear equations}, \textit{limit}, \textit{differentiation}, \textit{integral} and \textit{finite integral}. Table~\ref{table_datasets} displays the category wise statistics of these data sets. Figure~\ref{fig_sample_img1} 
shows few sample images and their descriptions of \texttt{Math-Exp-Syn} data set. 
\begin{table}
\addtolength{\tabcolsep}{-5.7pt}
\begin{center}
\begin{tabular}{|l|l|l l l l l l l|l|}\hline
\textbf{Data set} & \textbf{Division} & \multicolumn{7}{|l|}{\textbf{No. images}} & \textbf{Total}\\ \cline{3-9}
 &  & {\sc le} & {\sc ie} & {\sc ple} & {\sc lt} & {\sc di} & {\sc in} & {\sc fin} &\\ \hline
\texttt{Math-Exp-Syn} & Training & 41K & 43K & 43K & 44K & 39K & 40K & 36K & 299K \\ 
 & Validation & 5K & 5K & 5K & 6K & 4K &5K & 4K & 37K \\
 & Test & 5K & 5K & 5K & 6K & 4K &5K & 4K & 37K \\ \hline
\texttt{Math-Exp} & Test &1K &0.64K &0.05K &0.68K &0.06K &0.2K &0.1K &2.7K \\ \hline
\end{tabular}
\end{center}
\caption{Category level statistics of considered data sets. {\sc le}: linear equation, {\sc ie}: inequality, {\sc ple}: pair of linear equations, {\sc lt}: limit, {\sc di}: differentiation, {\sc in}: integral and {\sc fin}: finite integral.\label{table_datasets}}
\end{table}

\subsection{Evaluation Metrics}

In this work, we evaluate the generated descriptions for {\sc mei} with respect to three metrics: {\sc bleu}~\cite{papineni2002bleu}, {\sc cide}r~\cite{vedantam2015cider}, and {\sc rouge}~\cite{lin2004rouge} which are popularly used in natural language processing ({\sc nlp}) and image captioning tasks. All these metrics basically measure the similarity of a generated sentence against a set of ground truth sentences written by humans. Higher values of all these metrics indicate that the generated sentence (text) is more similar to the ground truth sentence (text).




\begin{figure*}[ht!]
\centerline{
\psfig{figure=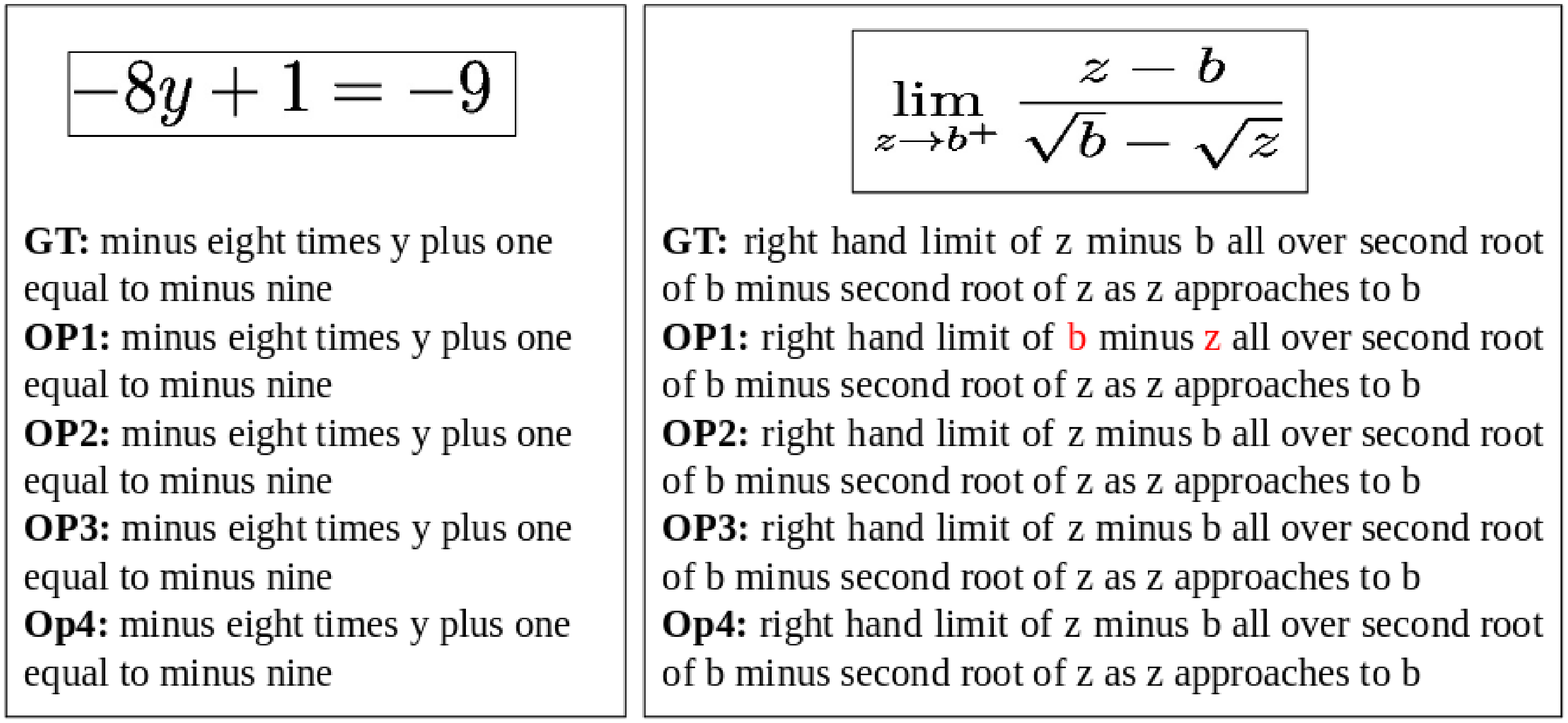,height=0.17\textwidth, width=0.5\textwidth}
\hspace{0.0001\textwidth}
\psfig{figure=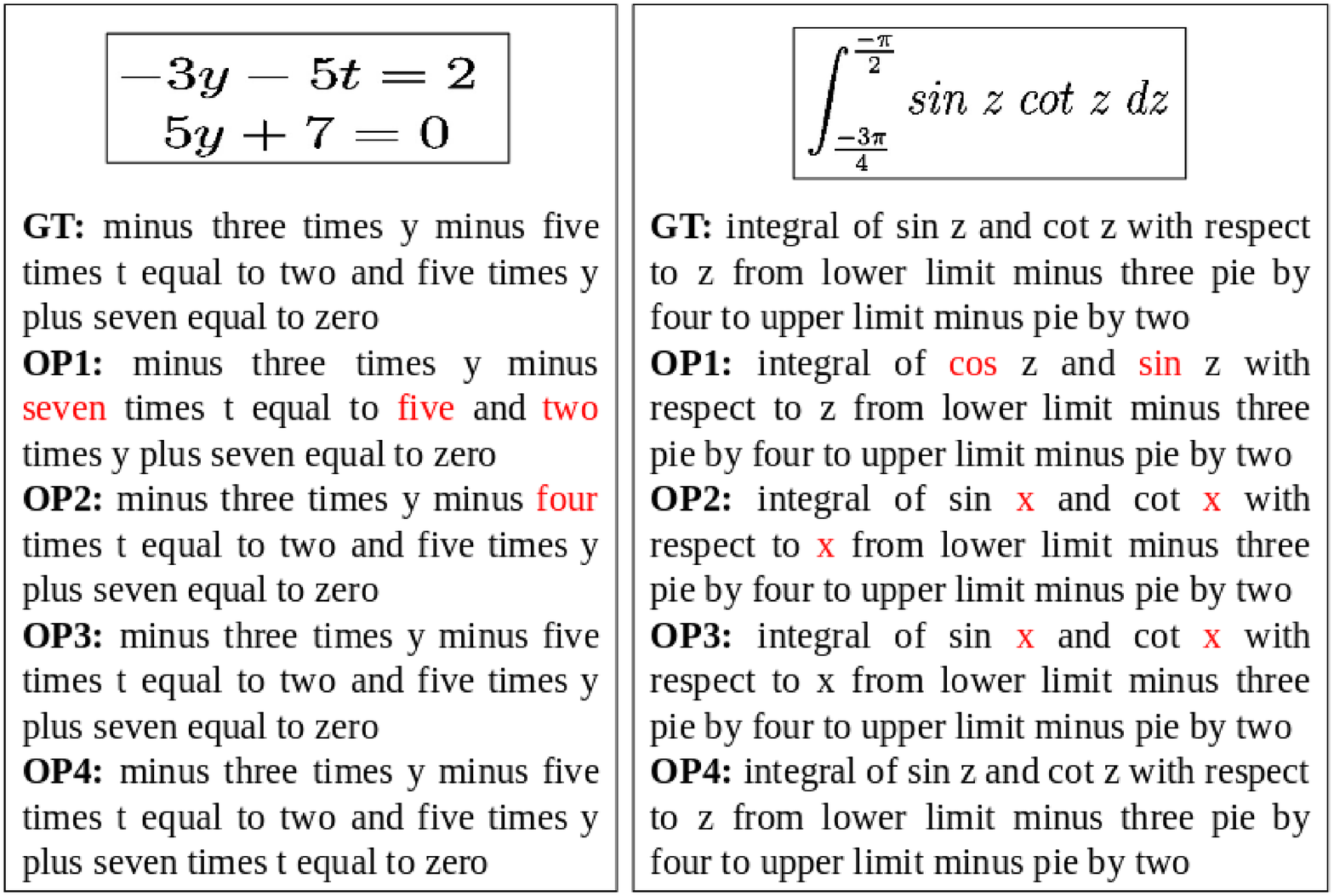,height=0.17\textwidth, width=0.5\textwidth}}
\vspace{0.01\textwidth}
\centerline{
\psfig{figure=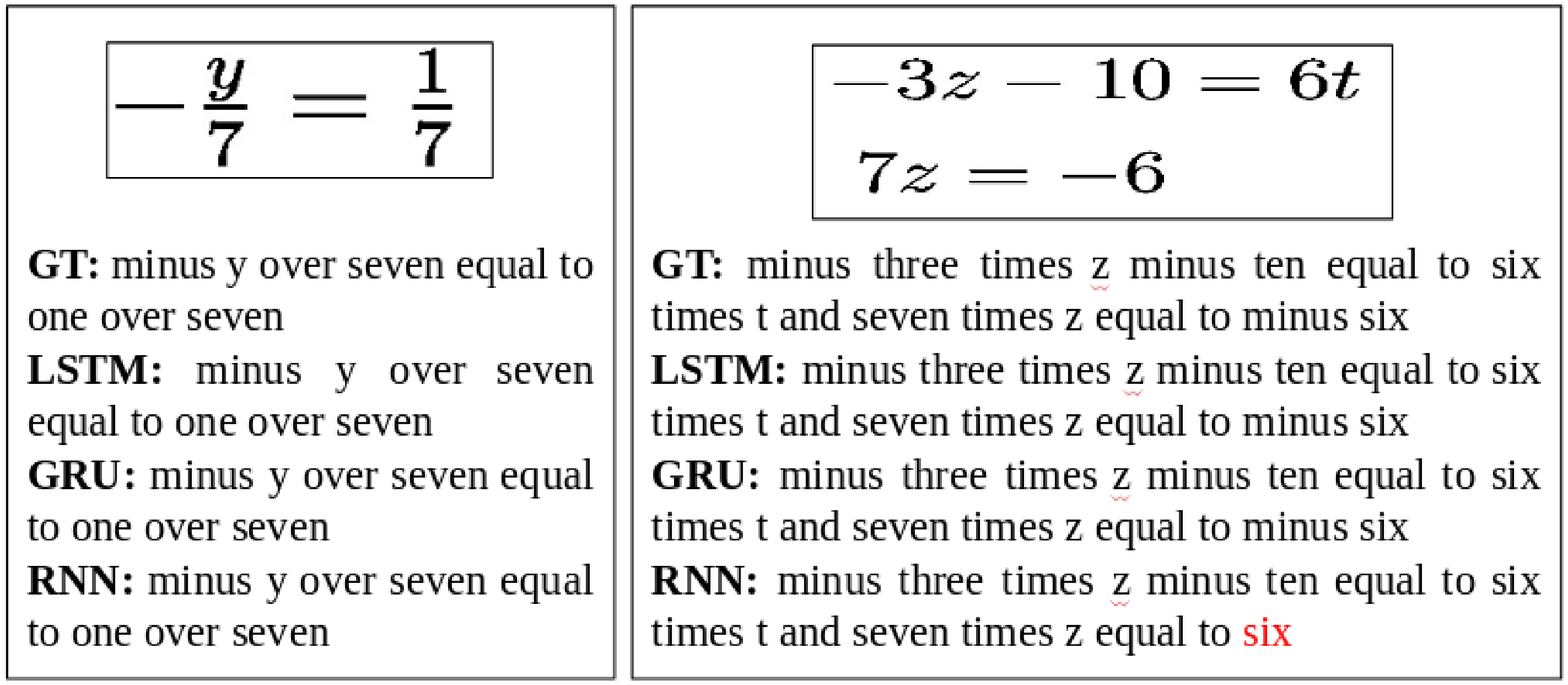,height=0.17\textwidth, width=0.5\textwidth}
\hspace{0.0001\textwidth}
\psfig{figure=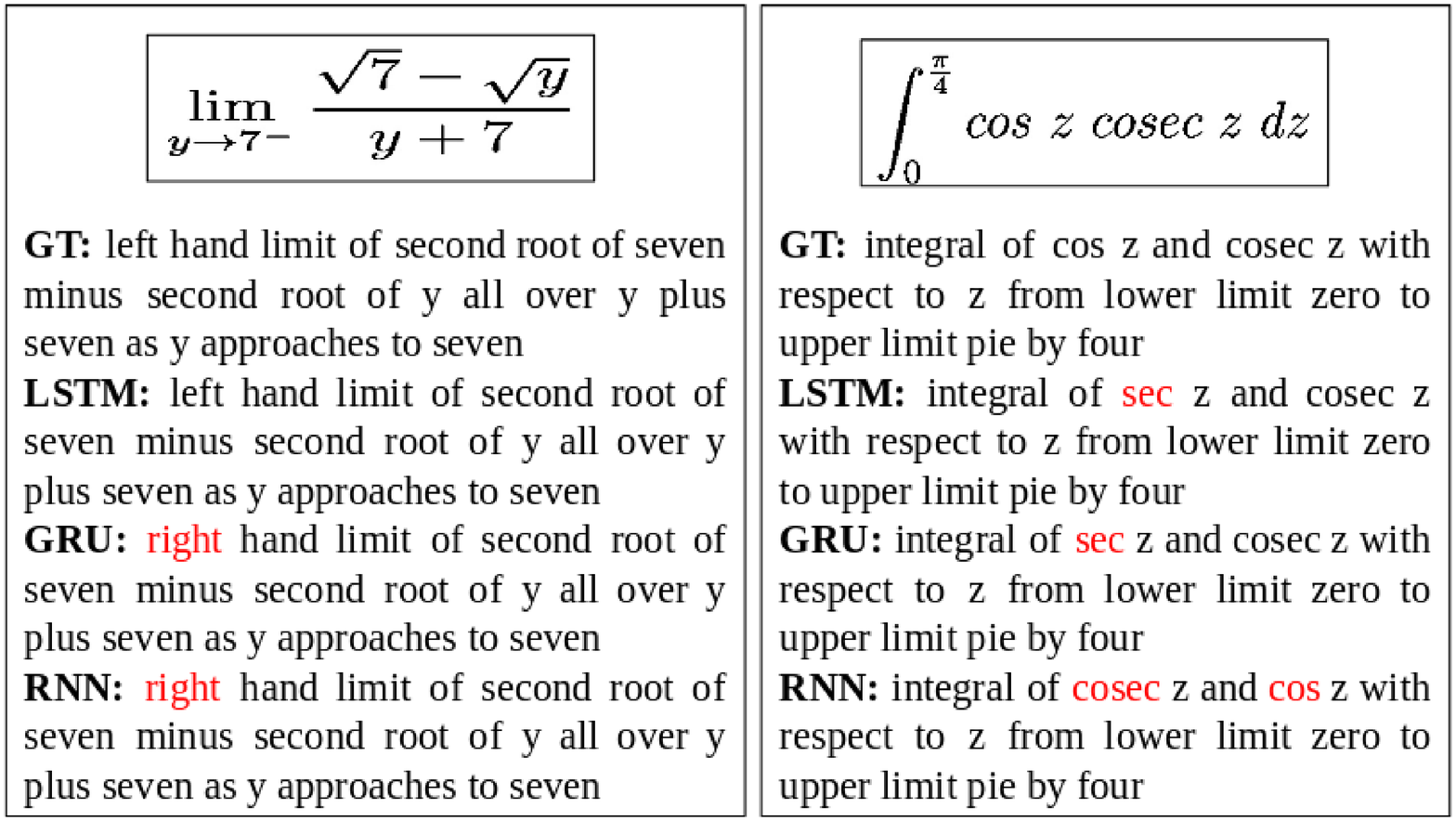,height=0.17\textwidth, width=0.5\textwidth}}
\vspace{0.01\textwidth}
\centerline{
\psfig{figure=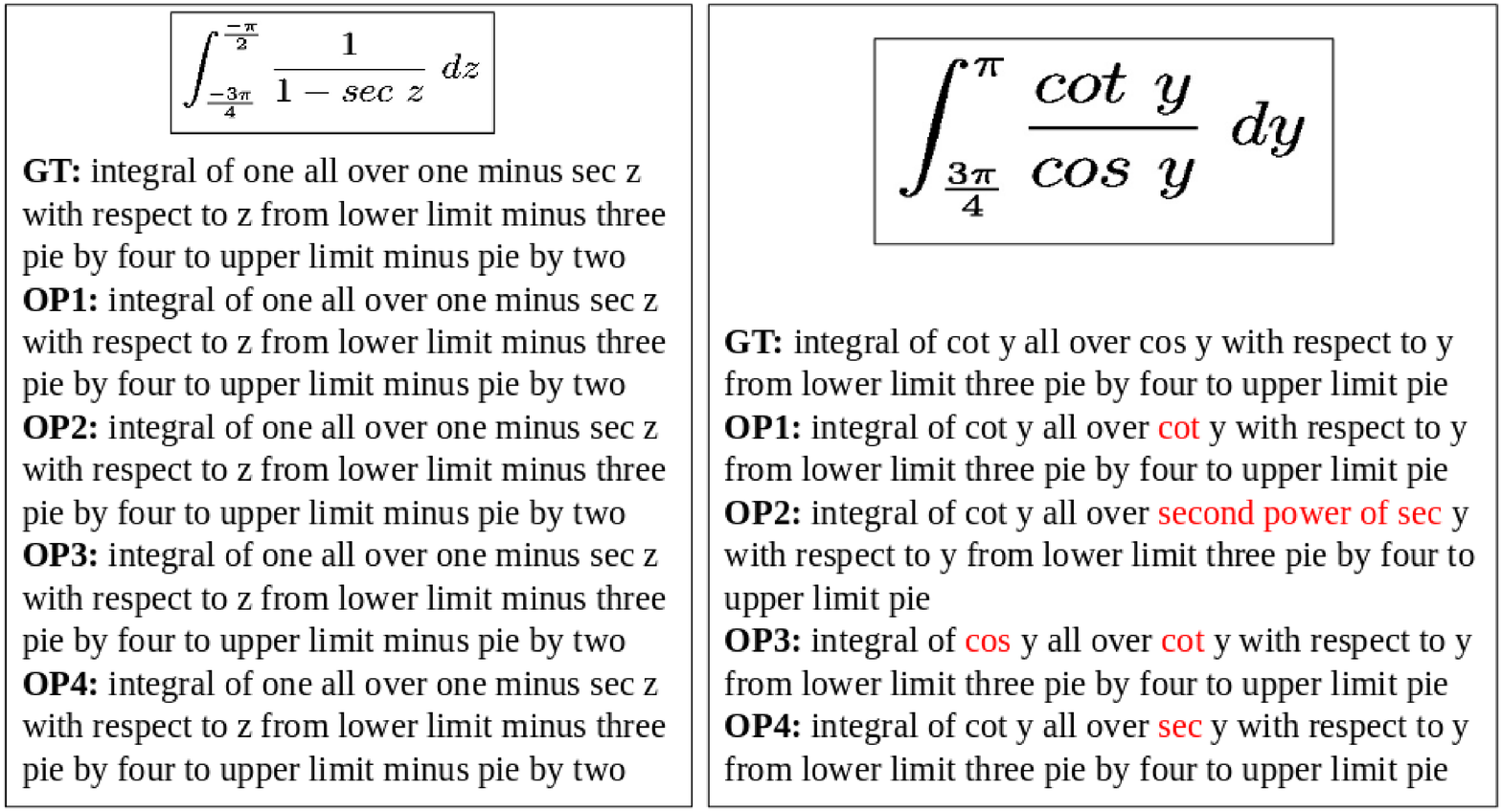,height=0.17\textwidth, width=0.5\textwidth}
\hspace{0.0001\textwidth}
\psfig{figure=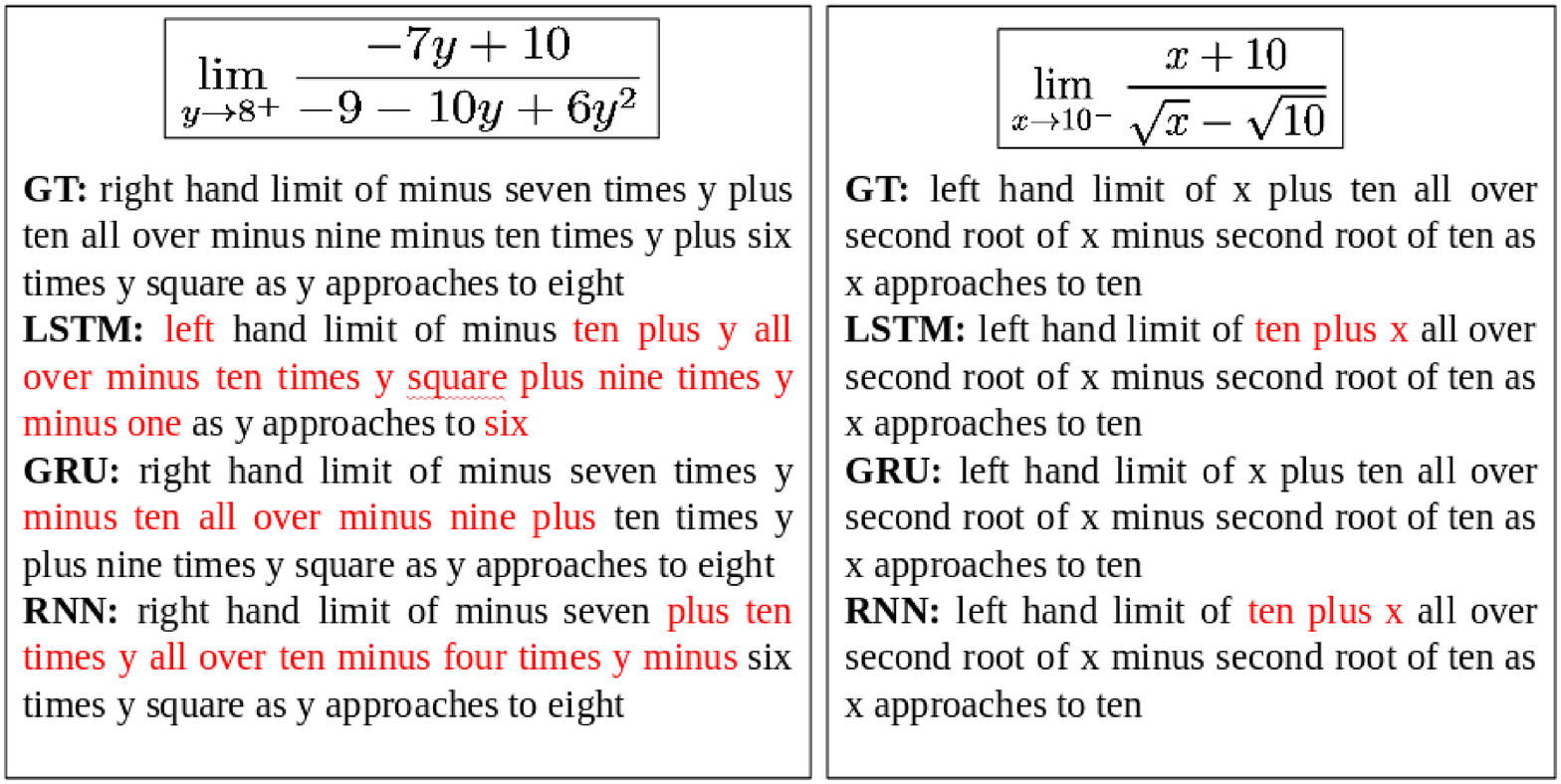,height=0.17\textwidth, width=0.5\textwidth}}
\caption{Visual illustration of sample results of test \texttt{Math-Exp-Syn} data set produced by our {\sc med} framework. {\sc \textbf{gt}}: ground truth description, {\sc \textbf{op1}}: description generated by ResNet152+{\sc lstm}, {\sc \textbf{op2}}: description generated by ResNet152+{\sc lstm}+Attn., {\sc \textbf{op3}}: description generated by ResNet152$^{\dagger}$+{\sc lstm} and {\sc \textbf{op4}}: description generated by ResNet152$^{\dagger}$+{\sc lstm}+Attn., 
{\sc \textbf{lstm}}: description generated by ResNet152$^{\dagger}$+{\sc lstm}+Attn., {\sc \textbf{gru}}: description generated by ResNet152$^{\dagger}$+{\sc gru}+Attn., {\sc \textbf{rnn}}: description generated by ResNet152$^{\dagger}$+{\sc rnn}+Attn., $^{\dagger}$ indicates fine-tune and Attn. denotes attention in decoder. Red colored text indicates wrongly generated text.  \label{fig_sample_result}}
\end{figure*}

\section{Experiments and Results}

An extensive set of experiments is performed to assess the effectiveness of our {\sc med} model using several metrics on the {\sc me} data sets.

\subsection{Ablation Study}

A number of ablation experiments is conducted to quantify the importance of each of the components of our algorithm and to justify various design issues in the context of mathematical equation description. \texttt{Math-Exp-Syn} data set is used for this purpose.

\paragraph{\textbf{Pre-trained Encoder}}
\begin{table}[ht!]
\addtolength{\tabcolsep}{-2.7pt}
\begin{center}
\begin{tabular}{|l| l l l l l l|} \hline
\textbf{Models} & \multicolumn{6}{ l| }{\textbf{Test Performance}} \\\cline{2-7} 
  & {\sc \textbf{b}-1} & {\sc \textbf{b}-2} & {\sc \textbf{b}-3} & {\sc \textbf{b}-4} & \textbf{c} & {\sc \textbf{r}} \\ \hline  
ResNet-18$^{\dagger}$+{\sc lstm}+Attn. &0.971 &0.949 &0.927 &0.906 &0.960 &9.014     \\
ResNet-34$^{\dagger}$+{\sc lstm}+Attn.  &0.973 &0.952 &0.931 &0.910 &0.962 &9.058 \\
ResNet-50$^{\dagger}$ +{\sc lstm}+Attn.  &0.978 &0.959 &0.940 &0.922 &0.968 &9.172   \\
ResNet-101$^{\dagger}$ +{\sc lstm}+Attn.  &0.979 &0.960 &\textbf{0.941} &\textbf{0.923} &0.968 &9.179   \\
ResNet-152$^{\dagger}$ +{\sc lstm}+Attn.  &\textbf{0.981} &\textbf{0.962} &\textbf{0.941} &\textbf{0.923} &\textbf{0.971} &\textbf{9.184} \\ \hline
\end{tabular}
\end{center}
\caption{It illustrates that the deeper pre-trained model gets better representation and improves textual description accuracy with respect to three evaluation measures: {\sc \textbf{bleu}}-1 ({\sc \textbf{b}}-1), {\sc \textbf{bleu}}-2 ({\sc \textbf{b}}-2), {\sc \textbf{bleu}}-3 ({\sc \textbf{b}}-3), {\sc \textbf{bleu}}-4 ({\sc \textbf{b}}-4), {\sc \textbf{cide}}\textbf{r} ({\sc \textbf{c}}) and {\sc \textbf{rouge}} ({\sc \textbf{r}}). Number along with the model refers to the depth of the corresponding model. `$^{\dagger}$' denotes that the encoder is fine-tuned during training. {\sc lstm} with attention is considered as a decoder.  \label{table_pretrained_model}} 
\end{table}
 
It is well known that the deeper networks are beneficial for the large scale image classification task. We conduct an experiment with different depths of the pre-trained models to analyze their performances on the mathematical equation description task. Detailed scores of equation description using the various pre-trained models are listed in Table~\ref{table_pretrained_model}.   

\paragraph{\textbf{Fine-tuned vs. Without Fine-tuned Encoder and Attention vs. Without Attention in Decoder}}

\begin{table}[ht!]
\addtolength{\tabcolsep}{-2.7pt}
\begin{center}
\begin{tabular}{|l|l l l l l l|} \hline
\textbf{Models} & \multicolumn{6}{ l| }{\textbf{Test Performance}} \\\cline{2-7} 
  & {\sc \textbf{b}-1} & {\sc \textbf{b}-2} & {\sc \textbf{b}-3} & {\sc \textbf{b}-4} & \textbf{c} & {\sc \textbf{r}} \\ \hline  
ResNet-152+{\sc lstm}. &0.976 &0.956 &0.937 &0.918 &0.965 &9.156   \\
ResNet-152+{\sc lstm}+Attn. &0.977 &0.956 &0.937 &0.918 &0.966 &9.163 \\  
ResNet-152$^{\dagger}$+{\sc lstm}. &0.978 &0.960 &0.940 &0.922 &0.968 &9.182   \\
ResNet-152$^{\dagger}$ +{\sc lstm}+Attn.  &\textbf{0.981} &\textbf{0.962} &\textbf{0.941} &\textbf{0.923} &\textbf{0.971} &\textbf{9.184}   \\  \hline \end{tabular}
\end{center}
\caption{Quantitative illustration of effectiveness of fine-tuning the encoder and attention in decoder during training on {\sc med} task. `$\dagger$' denotes fine-tune. \label{table_finetune_model}} 
\end{table}

The considered encoder, ResNet-$152$ pre-trained on ImageNet~\cite{imagenet_cvpr09} is not effective without fine-tuning due to domain heterogeneity (natural images and {\sc mei}s). We perform an experiment to establish potency of fine-tuning on the equation description task. The attention mechanism tells the decoder to focus on a particular region of the image while generating the description related to that region of the image. We do an experiment to analyze the effectiveness of attention mechanism on the mathematical equation description task. 
Observation of the experiments is quantitatively reported in Table~\ref{table_finetune_model}. This table highlights the effectiveness of fine-tune and attention in mathematical equation description task. First row of Figure~\ref{fig_sample_result} visually illustrates the effectiveness of fine-tuning the pre-trained ResNet-$152$ and {\sc lstm} with attention for {\sc med} task. 

\paragraph{{\sc \textbf{rnn}} \textbf{vs.} {\sc \textbf{gru}} \textbf{vs.} {\sc \textbf{lstm}}}

We also conduct an experiment to analyze performances of {\sc lstm}, Gated Recurrent Units ({\sc gru}) and Recurrent Neural Networks ({\sc rnn}) on generating captions for mathematical equation images. In this experiment, we consider pre-trained ResNet-$152$ as an encoder which is fine-tuned during training and different decoders: {\sc rnn}, {\sc gru} and {\sc lstm} with attention mechanism. Table~\ref{table_language_model} displays the numerical comparison between three decoder models. The table highlights that {\sc lstm} is more effective than other two models for mathematical equation description task. Second and third rows of Figures~\ref{fig_sample_result} display the visual outputs. This figure highlights that {\sc lstm} is able to generate text most similar to the ground truth.   
\begin{table}[ht!]
\addtolength{\tabcolsep}{-3.0pt}
\begin{center}
\begin{tabular}{|l |l l l l l l |} \hline
\textbf{Models} & \multicolumn{6}{ l |}{\textbf{Test Performance}} \\\cline{2-7} 
  & {\sc \textbf{b}-1} & {\sc \textbf{b}-2} & {\sc \textbf{b}-3} & {\sc \textbf{b}-4} & \textbf{c} & {\sc \textbf{r}} \\   \hline
ResNet-152$^{\dagger}$+{\sc rnn}+Attn.    &0.977 &0.958 &0.939 &0.920 &0.967 &9.179   \\
ResNet-152$^{\dagger}$+{\sc gru}+Attn.    &0.979 &0.959 &0.939 &0.920 &0.968 &9.182   \\ 
ResNet-152$^{\dagger}$ +{\sc lstm}+Attn.  &\textbf{0.981} &\textbf{0.962} &\textbf{0.941} &\textbf{0.923} &\textbf{0.971} &\textbf{9.184}   \\ \hline 
\end{tabular}
\end{center}
\caption{Performance comparison between {\sc rnn}, {\sc gru} and {\sc lstm} with attention mechanism on the mathematical equation description task. We fine-tune the encoder during training process.\label{table_language_model}} 
\end{table}

\begin{table}[ht!]
\addtolength{\tabcolsep}{-5.0pt}
\begin{center}
\begin{tabular}{|l | l | l|  l l l l l l |} \hline
\textbf{Models} & \textbf{Data sets} & \textbf{Division} & \multicolumn{6}{ l| }{\textbf{Scores}} \\\cline{4-9} 
 & & & {\sc \textbf{b}-1} & {\sc \textbf{b}-2} & {\sc \textbf{b}-3} & {\sc \textbf{b}-4} & \textbf{c} & {\sc \textbf{r}} \\ \hline   
{\sc med} & \texttt{Math-Exp-Syn} & test set  &0.981 &0.962 &0.941 &0.923 &0.971 &9.184    \\ \hline  
{\sc med} & \texttt{Math-Exp} & test set  &0.975 &0.956 &0.936 &0.917 &0.966 &9.146 \\ \hline
\end{tabular}
\end{center}
\caption{Quantitative results of our {\sc med} model on standard evaluation metrics for both \texttt{Math-Exp-Syn} and \texttt{Math-Exp} data sets. Both the cases {\sc med} is trained using training set of \texttt{Math-Exp-Syn} data set. \label{table_quantitative_result}} 
\end{table}

\subsection{Quantitative Analysis of Results}

The quantitative results obtained using our {\sc med} model for both \texttt{Math-Exp-Syn} and \texttt{Math-Exp} data sets are listed in Table~\ref{table_quantitative_result}.  

\begin{figure}
\centerline{
\psfig{figure=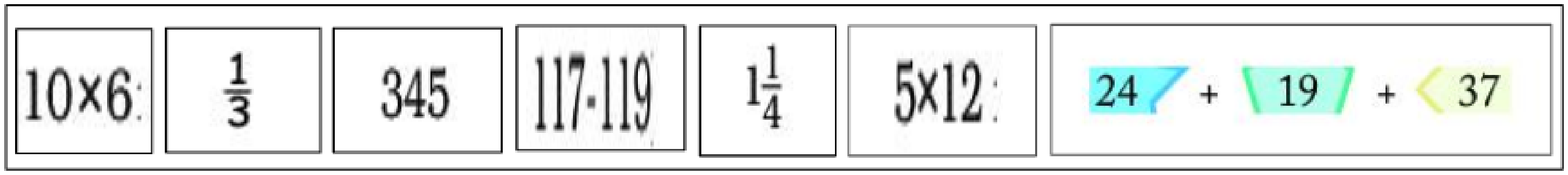,height=0.07\textwidth, width=0.5\textwidth}}
\caption{Sample cropped mathematical equation images from \textsc{ncrt} class V mathematical book for real world experiment.\label{fig_ncrt}}
\end{figure}

\begin{table}[ht!]
\addtolength{\tabcolsep}{-5.7pt}
\begin{center}
\begin{tabular}{|l|l|l|} \hline
\textbf{Cropped } & \textbf{Textual description } & \textbf{Equation Written}  \\ 
 \textbf{image} & \textbf{generated by \textsc{med}} & \textbf{by Students}   \\ \hline 
\psfig{figure=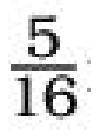,height=0.03\textwidth, width=0.08\textwidth} & five by sixteen & $\frac{5}{16}$\\ \hline
\psfig{figure=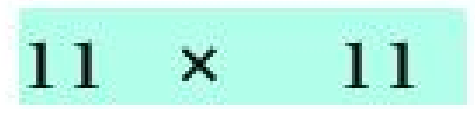,height=0.03\textwidth, width=0.08\textwidth} & eleven into eleven & $11 \times 11$ \\ \hline
\psfig{figure=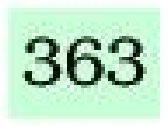,height=0.03\textwidth, width=0.08\textwidth} & three hundred and sixty three  & $363$\\ \hline
\psfig{figure=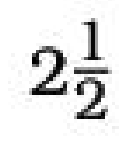,height=0.03\textwidth, width=0.08\textwidth} & two and one by two & $2\frac{1}{2}$\\ \hline
\psfig{figure=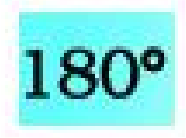,height=0.03\textwidth, width=0.08\textwidth} & one hundred and eighty degree & $180^{o}$\\ \hline
\psfig{figure=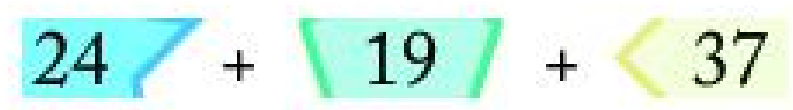,height=0.03\textwidth, width=0.08\textwidth} & eight plus nine plus three & $8+9+3$\\ \hline
\psfig{figure=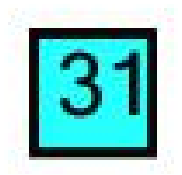,height=0.03\textwidth, width=0.08\textwidth} & one thousand three hundred and eleven &$1131$\\ 
\hline
\psfig{figure=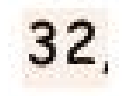,height=0.03\textwidth, width=0.08\textwidth} & thirty two point & $32.$\\ \hline
\end{tabular}
\end{center}
\caption{Summary of real world experiments. \textbf{First Column:} cropped equation images. \textbf{Second Column:} textual descriptions generated by the \textsc{med} and given to the students and ask them to write corresponding equations by reading the descriptions. \textbf{Third Column:} equations written by the students. \label{table_realworld_experiment}} \end{table}

\subsection{Real world Experiments}

We conduct a real world experiment to see whether the students are able to write the equations by only reading or listening their textual descriptions or not. For this purpose, we create a test set of mathematical equation images which are cropped from \textsc{ncrt} class V mathematical book\footnote{\url{https://www.ncertbooks.guru/ncert-maths-books/}}. Test set consists of $398$ cropped equation images of various types of equations: integer, decimal, fraction, addition, subtraction, multiplication and division. Figure~\ref{fig_ncrt} shows the sample cropped mathematical equation images from \textsc{ncrt} class V mathematical book. Our {\sc med} system generates the textual description for each of these equations. The list of descriptions of equations is given to the students and ask to write the corresponding mathematical equations within $1$ hour. Twenty students participate in this test. If anyone of the students writes the incorrect equations by only reading or listening their textual descriptions. Then the answer is wrong otherwise correct. Among $398$ equations, students are able to correctly write $359$ equations within time by reading their textual descriptions generated by our \textsc{med} model. For remaining $39$ equations, our \textsc{med} model generates wrong descriptions due to the presence of other structural elements (i.e. triangle, square, etc). Table~\ref{table_realworld_experiment} highlights the few results of this test. Since, descriptions generated by \textsc{med} model are wrong, students write wrong equations by reading wrongly generated descriptions. From this test, we conclude that our {\sc med} model is effective for reading equations by generating their textual descriptions.

\section{Conclusions}

In this paper, we introduce a novel mathematical equation description ({\sc med}) model for reading mathematical equations for blind and visually impaired students by generating textual descriptions of the equations. Unavailability of mathematical images and their textual descriptions, inspires us to generate two data sets for experiments. Real-world experiment concludes that the students are able to write mathematical expression by reading or listening their descriptions generated by the \textsc{med} network. This experiment establishes the effectiveness of the \textsc{med} framework for reading mathematical equation for the blind and \textsc{vi} students. 

\def\IEEEbibitemsep{0pt plus .5pt}
\bibliographystyle{IEEEtran}
\bibliography{egbib}

\end{document}